\documentclass[a4paper]{article}

\usepackage[margin=25mm]{geometry}
\usepackage{amsmath}
\usepackage{amsfonts}
\usepackage{amssymb}
\usepackage{graphicx}
\usepackage{verbatim}
\usepackage{authblk}
\usepackage{multirow}
\usepackage{textcomp}
\usepackage{multicol}
\usepackage{subfigure}
\usepackage{appendix}
\usepackage{float}
\usepackage[export]{adjustbox}
\usepackage[table,xcdraw]{xcolor}

\usepackage{caption}

\providecommand{\keywords}[1]
{
  \small	
  \textbf{\textit{Keywords---}} #1
}

\title{Neuronal Cell Type Classification using Deep Learning}

\author[1, 2]{Ofek Ophir}
\author[1, *]{Orit Shefi}
\author[1, *]{Ofir Lindenbaum}
\affil[1]{Faculty of Engineering, Bar-Ilan University, Ramat-Gan 5290002, Israel}
\affil[2]{The Leslie \& Susan Goldschmied (Gonda) Multidisciplinary Brain Research Center, Bar-Ilan University, Ramat-Gan 5290002, Israel}
\affil[*]{Corresponding author}

\begin{document}

\maketitle

\begin{abstract}
The brain is likely the most complex organ, given the variety of functions it controls, the number of cells it comprises, and their corresponding diversity. Studying and identifying neurons, the brain's primary building blocks, is a crucial milestone and essential for understanding brain function in health and disease. Recent developments in machine learning have provided advanced abilities for classifying neurons. However, these methods remain black boxes with no explainability and reasoning. This paper aims to provide a robust and explainable deep-learning framework to classify neurons based on their electrophysiological activity. Our analysis is performed on data provided by the Allen Cell Types database containing a survey of biological features derived from single-cell recordings of mice and humans. First, we classify neuronal cell types of mice data to identify excitatory and inhibitory neurons. Then, neurons are categorized to their broad types in humans using domain adaptation from mice data. Lastly, neurons are classified into sub-types based on transgenic mouse lines using deep neural networks in an explainable fashion. We show state-of-the-art results in a dendrite-type classification of excitatory vs. inhibitory neurons and transgenic mouse lines classification. The model is also inherently interpretable, revealing the correlations between neuronal types and their electrophysiological properties.
\end{abstract}


\section{Introduction}\label{intro}
The brain is a highly complex system with billions of neurons propagating signals to communicate and share information. Proper functionality of the nervous system requires mechanisms for information sharing between many neurons in different brain regions. Understanding these mechanisms remains an open and challenging problem in biology and requires a detailed and exact description of all brain regions and the neurons composing them. \par

The task of classifying neurons, the building blocks of the nervous system, has been an ongoing challenge in neuroscience ever since Ramon y Cajal's 'Histology of the Nervous System of Man and Vertebrates' \cite{cajal1995histology} was published, which was to a certain degree, an attempt to classify neurons. Neuroscientists attempting to study the nervous system have hypothesized that the differences in neuron morphology play a role in the neural circuit. For this reason, it is essential to accurately classify the different types of neurons \cite{zeng2017neuronal}. \par

Defining a solid neuronal cell-type taxonomy is challenging and includes two significant obstacles. First, classification studies were underpowered and laborious, which caused highly biased results. However, in the past decade, technological advances have made it possible to analyze hundreds of neurons accurately and efficiently \cite{zhao2008patch}. The second obstacle is determining how fine and firm the distinctions between neuronal types should be. If the resolution is too broad (such as the distinction between sensory and motor neurons), the taxonomy might be too coarse and have little value for experimental purposes. However, if the resolution is too fine, the neuronal taxonomy might have no relevance (an extreme case would be to think of each neuron as an independent type). \par

In this paper, we carry out classification using electrophysiological features such as action potential (AP) threshold, width, height, hyperpolarization voltage, and resting potential. These features are aimed to describe the differences among the observed variability in neuronal activities and can be used to define electrophysiological types of neurons \cite{beierlein2003two, nowak2008lack}. The axonal morphologies also impact AP propagation and are relevant in neurons' complex axonal ramification patterns \cite{ofer2020axonal, ofer2016axonal}. It should be noted that electrophysiological features are more accessible to measure than morphological or genetic features and can be simultaneously recorded using techniques such as optical imaging of electrical activity on hundreds of neurons \cite{zeng2017neuronal}.

We present a deep learning framework for predicting neuronal types using electrophysiological features in two classification tasks. The first is classifying neurons to their broad type, excitatory vs. inhibitory. The second task includes classifying neurons into their inhibitory subclasses and excitatory neurons into their broad class in mice. For both classification tasks, we use data from the Allen Cell Types Database - ACTB \cite{allen2015allen}, a publicly available brain cell database containing recordings of electrical stimulus and response in different types of neurons from both human and mouse cells. \par

Since obtaining neuronal data from humans is challenging, we use additional data from mice to develop a reliable classifier that exploits information from humans and mice. We present a domain adaptive neural network that classifies broad neuronal cell types from both domains. Next, we use mice data to identify granular neuronal subclasses and implement a locally sparse network that enables reliable classification in a low sample size regime. The model leads to state-of-the-art results compared with leading baselines while providing high interpretability, which is vital for future research.


\section{Background}\label{background}
At the most fundamental level, cells can be classified into non-neuronal cells and neurons, which can then be further classified into excitatory and inhibitory neurons \cite{melzer2020diversity}. The significant difference between the two types is that the excitatory neurons release neurotransmitters (most commonly glutamic acid) that fire an action potential in the postsynaptic neuron. In contrast, inhibitory neurons release neurotransmitters (most commonly gamma-aminobutyric acid - GABA) which inhibit the firing of an action potential. Inhibitory interneurons comprise only 10-20\% of the total neural population in the cortex but are essential for sensation, movement, and cognition \cite{swanson2019hiring}. \par

Excitatory neurons are usually morphologically spiny, with a long apical dendrite, and exhibit less variability in their electrophysiological features. This makes it harder to distinguish between excitatory cell types solely using electrophysiological characteristics. Inhibitory neurons are typically aspiny or sparsely spiny, with a more compact dendritic structure, having a more considerable variance in electrophysiological properties and tending to spike faster \cite{kawaguchi1997neostriatal, strubing1995differentiation}. Neurons can also be classified based on their neurotransmitter, GABAergic neurons which are mostly inhibitory cells, and Glutamatergic neurons, which are habitually excitatory and brain-area specific. \par

Many neuroscientists also consider GABAergic neurons as belonging to one of the four subclasses based on the expression of specific principal markers; these include: 
\begin{itemize}
  \item Pvalb (parvalbumin) positive. 
  \item Vip (vasoactive intestinal peptide) positive.
  \item Sst (somatostatin) positive.
  \item Htr3a (5-hydroxytryptamine receptor) positive, Vip negative.
\end{itemize}
These subclasses of GABAergic interneurons account for most neurons in specific brain regions. The classes are expressed in a non-overlapping manner, meaning that each neuron belongs to one class in a monovalent fashion, with distinct cell types accompanied by different physiological properties \cite{tremblay2016gabaergic}. \par

\begin{figure}[ht]
\centering
\includegraphics[width=0.85\textwidth]{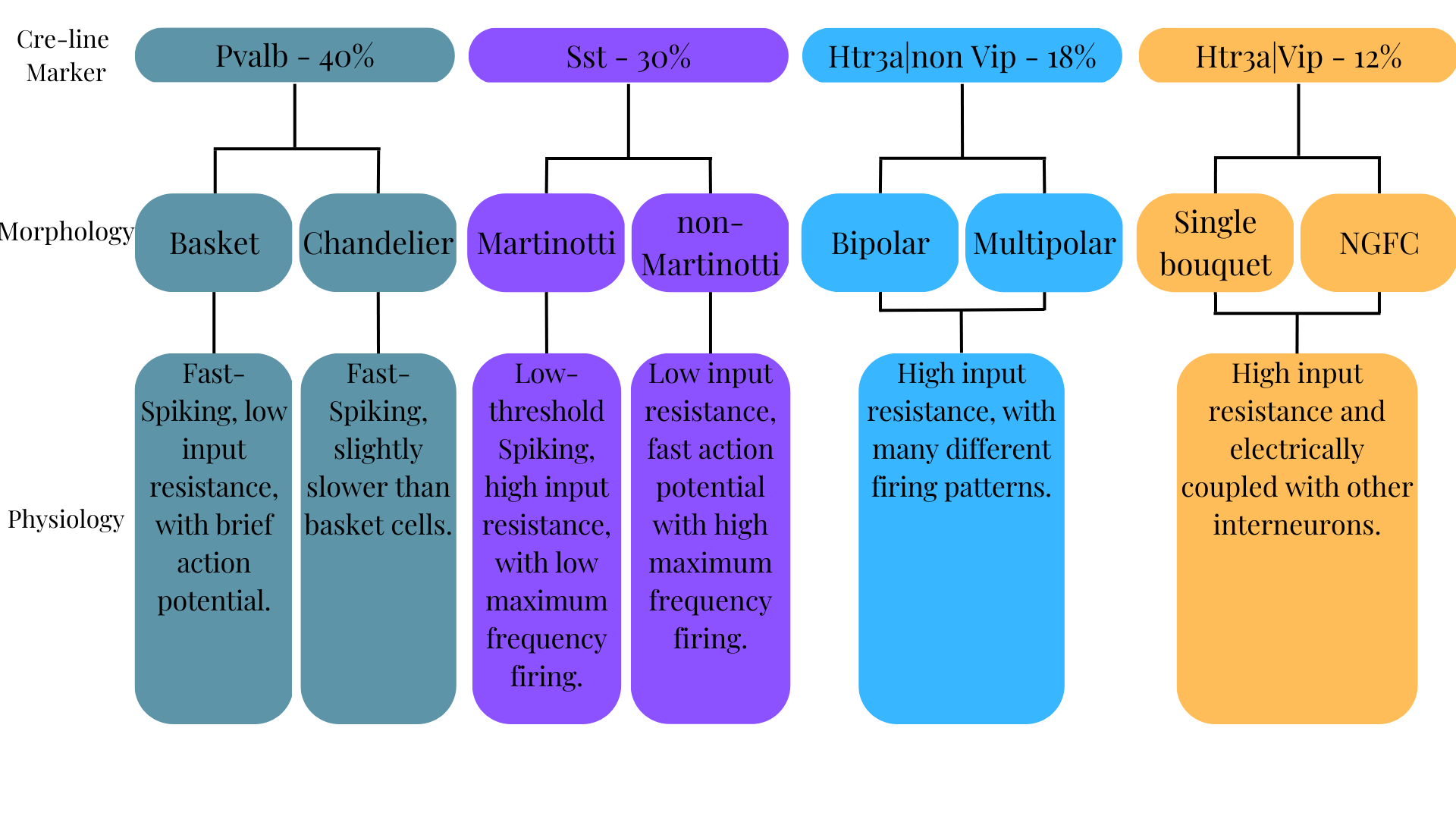}
\caption{GABAergic neurons in the neocortex express one of four markers: Pvalb (Parvalbumin), Sst (Somatostatin), and the ionotropic serotonin receptor Ht3ar which can be either Vip (Vasoactive intestinal polypeptide) positive or negative.}
\label{fig1}
\end{figure}


\section{Related Work}\label{previous work}
In 2019, the ACTB became public \cite{allen2015allen, gouwens2019classification}, and with recent advances in computing capabilities and rapid development of machine and deep learning methods, the domain of neuronal cell-type classification has leaped forward. From the ACTB, 17 electrophysiological neuron types were identified, 4 of which were classified as excitatory subtypes, and 13 were inhibitory. The 13 inhibitory subtypes were further mapped into the four inhibitory interneuron types based on genetic tags: Vip, Ndnf, Sst, and Pvalb. The researchers also identified 38 morphological, and 46 morpho-electric neuron types, all of which were classified using current clamp electrophysiological recordings and the help of dimensionality reduction algorithms such as principal component analysis \cite{abdi2010principal} and t-distributed stochastic neighbor embedding \cite{van2008visualizing}. \par

Using the ACTB, Ghaderi et al., \cite{ghaderi2018electrophysiological} developed a semi-supervised method in which neuron classification occurs within three types of neurons. These types are excitatory pyramidal cells (Pyr), parvalbumin-positive (Pvalb) interneurons, and somatostatin-positive (Sst) interneurons from layer 2/3 of the mouse primary visual cortex. The authors achieved accuracies of 91.59 ± 1.69, 97.47 ± 0.67, and 89.06 ± 1.99 for Pvalb, Pyr, and Sst, respectively, which yielded an overall accuracy of 92.67 ± 0.54\%. \par

In 2019, Seo et al. \cite{seo2019predicting} used machine learning to predict transgenic markers of neurons using electrophysiological recordings. The work evaluated three different methods, namely: random forest - RF \cite{breiman2001random}, least absolute shrinkage and selection operator – LASSO \cite{tibshirani1996regression}, and artificial neural networks - ANN \cite{rosenblatt1958perceptron}. The prediction performance of the three models was similar, with 28.57-46.93\% accuracy in predicting the transgenic marker of excitatory neurons (Ctgf, Cux2\&Slc17, Nr5a1\&Scnn1a, Ntsr1, Rbp4, and Rorb) and 59.03-73.49\% accuracy at predicting the transgenic marker of inhibitory neurons (Chrna2, Gad2, Htr3a, Ndnf, Nkx2, Pvalb, Sst, Vip\&Chat). \par

In 2021, Rodríguez et al. \cite{rodriguez2021electrophysiological} revealed a circular ordered taxonomy using a transformation of the first two principal components and validated the proposed taxonomy with machine learning models (linear discriminant analysis – LDA \cite{balakrishnama1998linear}, RF, gradient boosted decision tree – GBDT \cite{natekin2013gradient}, support vector machine – SVM \cite{cortes1995support}, and ANN ensemble \cite{zhou2002neural}). These models were able to discriminate the different neuron types (4 types of inhibitory neurons – Pvalb, Htr3a, Sst, Vip, as well as Glutamatergic excitatory cells) using electrophysiological features with accuracy ranging between 66.1-75.2\% for the raw data, and 72.0-80.3\% accuracy for a subset of the data that has been cleaned using anomaly detectors. \par

It is worth noting that these studies only used mouse data, which is most likely due to insufficient human data. One can train a machine-learning model on data from humans and mice to overcome this limitation. However, training a model on multiple domains may result in overfitting to that most abundant domain, and lead to a performance gap data from other domains \cite{novak2018sensitivity, farahani2021brief}. This is known as the domain shift problem and can be addressed using tools from domain adaptation. Another issue with using these types of algorithms emerges from the complexity of the neural networks, making it difficult to interpret the model’s decisions \cite{gunning2019xai}. Model interpretability is essential to bio-medicine, where practitioners must trust the machine learning model. \par

This paper addresses these issues by providing a machine-learning framework for predicting neuronal cell types in two steps. The first is classifying excitatory vs. inhibitory neurons, and the second is classifying excitatory Glutamatergic cells and the different subclasses (Pvalb, Htr3a, Sst, and Vip) of inhibitory GABAergic cells. First, regarding dendrite type classification, we use mus-musculus (house mouse) source data which we have in a larger quantity, to learn a distribution over the homo sapiens (human) target data using domain adaptation methods. By doing this, we improve the results and robustness of the model on the target data. We also explain the importance of the unique features during testing using a concept from cooperative game theory. Then, we use a deep neural network inherently explainable model to predict Cre-line labels from mouse data; these account for the different GABAergic neuron subclasses and Glutamatergic neurons.

\pagebreak

\section{Data}\label{data}
\subsection{Allen Cell Types Database}\label{allencell}
The ACTB contains electrophysiological recordings from 1920 mice and 413 human cells.
The cells from humans and mice are categorized by dendrite type: spiny, aspiny, and sparsely spiny, as well as location and layer in the brain. The cells from mice samples are further identified for isolation using transgenic mouse lines harboring fluorescent reporters, with drivers that allow enrichment for cell classes based on marker genes such as Pvalb positive, Sst positive, Vip positive, and Htr3a (5-hydroxytryptamine receptor) positive but Vip negative \cite{tremblay2016gabaergic}. \par

The mouse data contains whole-cell current clamp recordings from identified fluorescent Cre-positive neurons or nearby Cre-negative neurons in acute brain slices derived from adult mice. The human data contains whole-cell current clamp recordings from adult human neocortical neurons in brain slices derived from surgical specimens.
Each whole-cell current clamp recording is a response to a stimulation recorded at 200 KHz (before 2016) or 50 KHz (after 2016).
Electrophysiological features are calculated and extracted from each whole-cell current clamp recording according to Table \ref{table:features}. \par

The stimulations types include:
\begin{itemize}
  \item Noise - Noise pulses offset with square current injections.
  \item Ramp - a current injection of increasing intensity at a rate slower than the time constant of the neuron.
  \item Long square - a square pulse of duration that allows the neuron to reach a steady state.
  \item Short square - a square pulse brief enough to elicit a single action potential.
\end{itemize} 

\begin{figure}[h]
\centering
\includegraphics[width=0.9\textwidth]{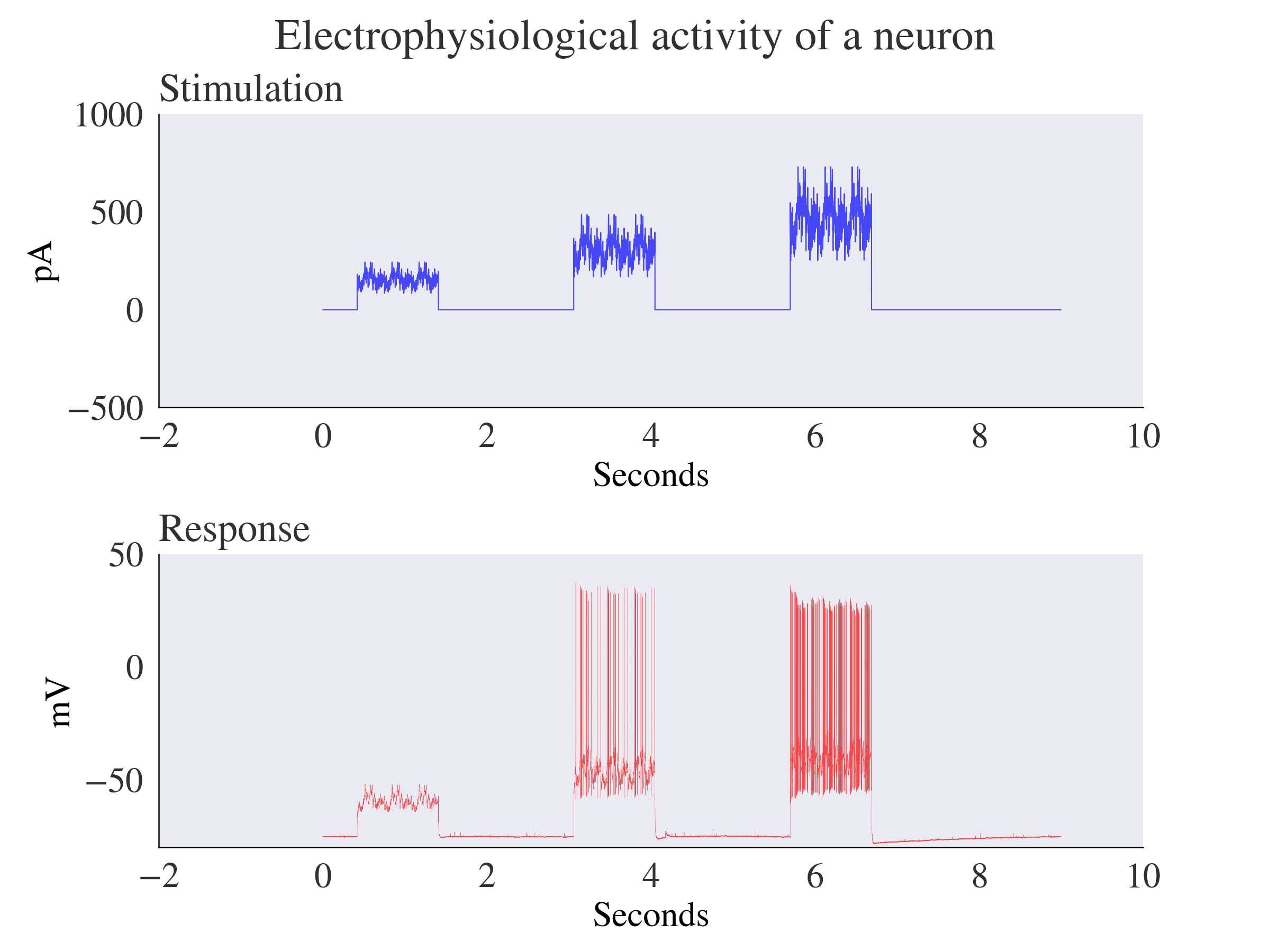}
\caption{(Top) Stimulation of noise pulses with square current injections scaled to three amplitudes, 0.75, 1, and 1.5 times Rheobase with a coefficient of variation (CV) equal to 0.2, as it resembles in vivo data. (Bottom) cell response to the noise stimulation.}
\label{fig:stim_resp}
\end{figure} \par
\subsection{Pre-processing}\label{preprocessing}
\begin{figure}[ht]
\centering
\includegraphics[width=0.9\textwidth]{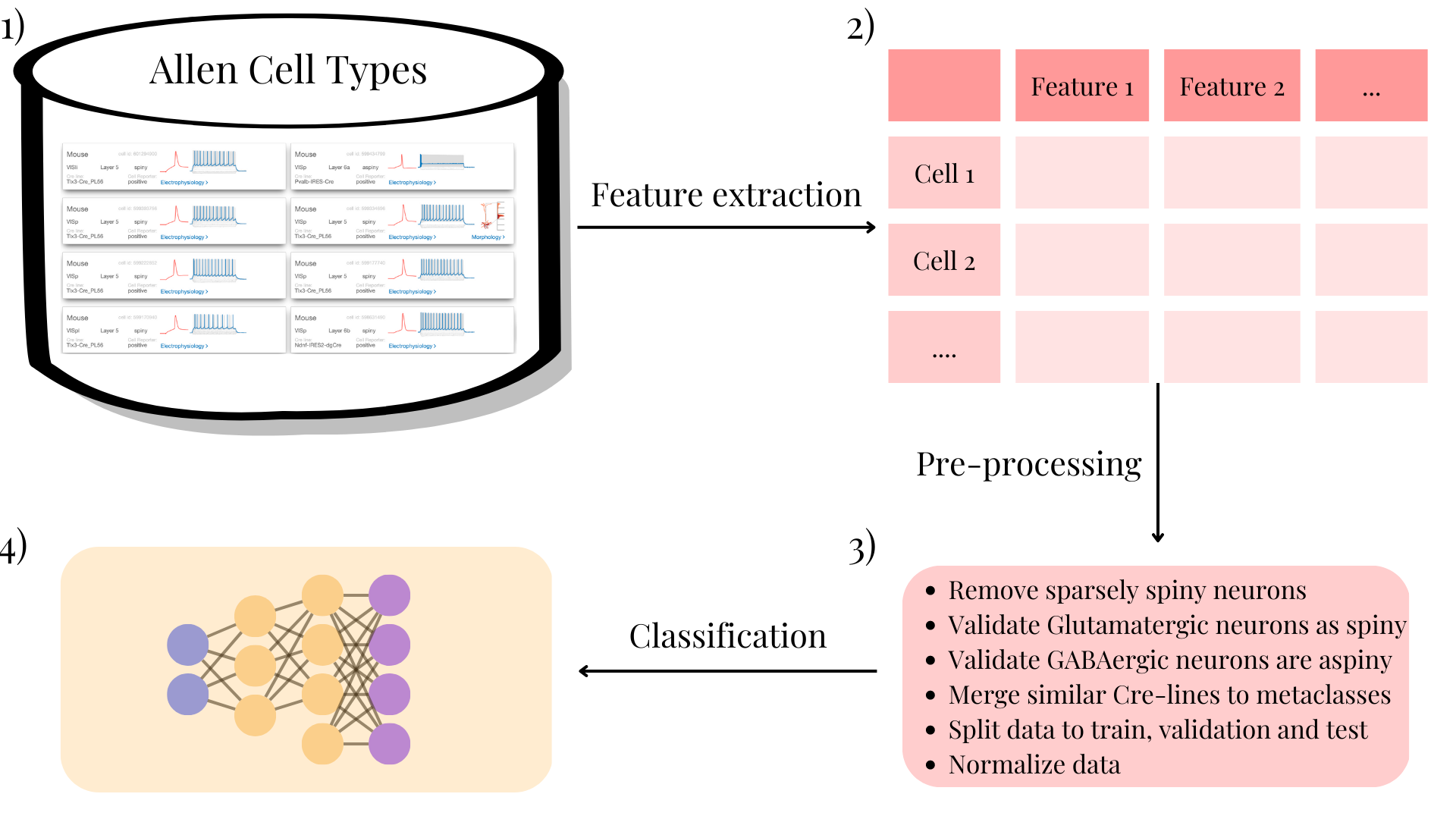}
\caption{Electrophysiological features pipeline. Data is first obtained from the ACTB; then, features are extracted into a tabular format. Similar Cre-lines are merged, and data is split into train, validation, and test. The data is normalized, and classification occurs through the neural network models.}
\label{fig:data_pipe}
\end{figure}

The data was downloaded from the ACTB. From each whole-cell patch clamp recording, electrophysiological features were extracted into tabular format, then, similar Cre lines were grouped. The entire process is illustrated in Figure \ref{fig:data_pipe}. \par

 \par

\begin{figure}[ht]
\centering
\includegraphics[width=0.8\textwidth]{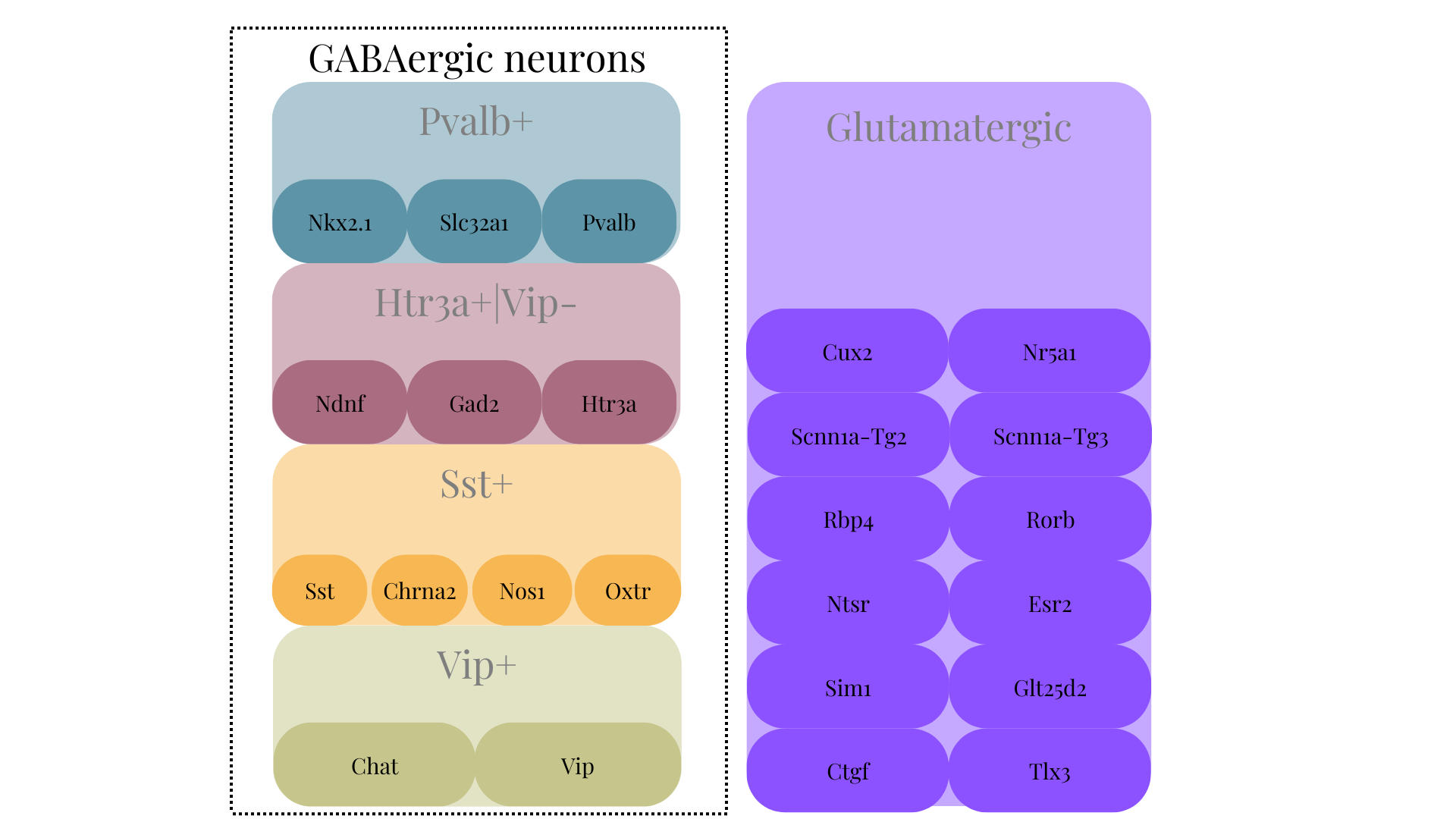}
\caption{Cre-lines composing the defined subclasses.}
\label{fig:subclasses}
\end{figure}

GABAergic neurons belong to four subclasses based on their expressed Cre lines, Pvalb (Parvalbumin) positive, Vip (Vasoactive intestinal peptide) positive, Sst (Somatostin) positive, and 5-hydroxytryptamine receptor 3A (Htr3a) positive Vip negative. Glutamatergic neurons belong to different subclasses according to their laminar locations and the location to which they project their axons \cite{tremblay2016gabaergic}. Using the ACTB, five transcriptomic-electrophysiological subclasses have been defined; these include four major GABAergic subclasses and one Glutamatergic subclass specified in Figure \ref{fig:subclasses}, according to \cite{rodriguez2021electrophysiological}. \par

 \par

\pagebreak
The mouse data is quite balanced regarding dendrite type distribution, with 700 spiny neurons (Glutamatergic) and 724 aspiny neurons (GABAergic). The human data contains more spiny neurons (231) than aspiny neurons (68). In terms of Cre-line subclasses within the mouse data, Glutamatergic neurons are the largest group with 700 neurons, Pvalb contains 231 neurons, Htr3a positive Vip negative contains 199 neurons, Sst contains 173 neurons, and Vip positive contains 121 neurons. \par

\begin{figure}[ht]
\centering
\subfigure{\includegraphics[width=0.49\textwidth]{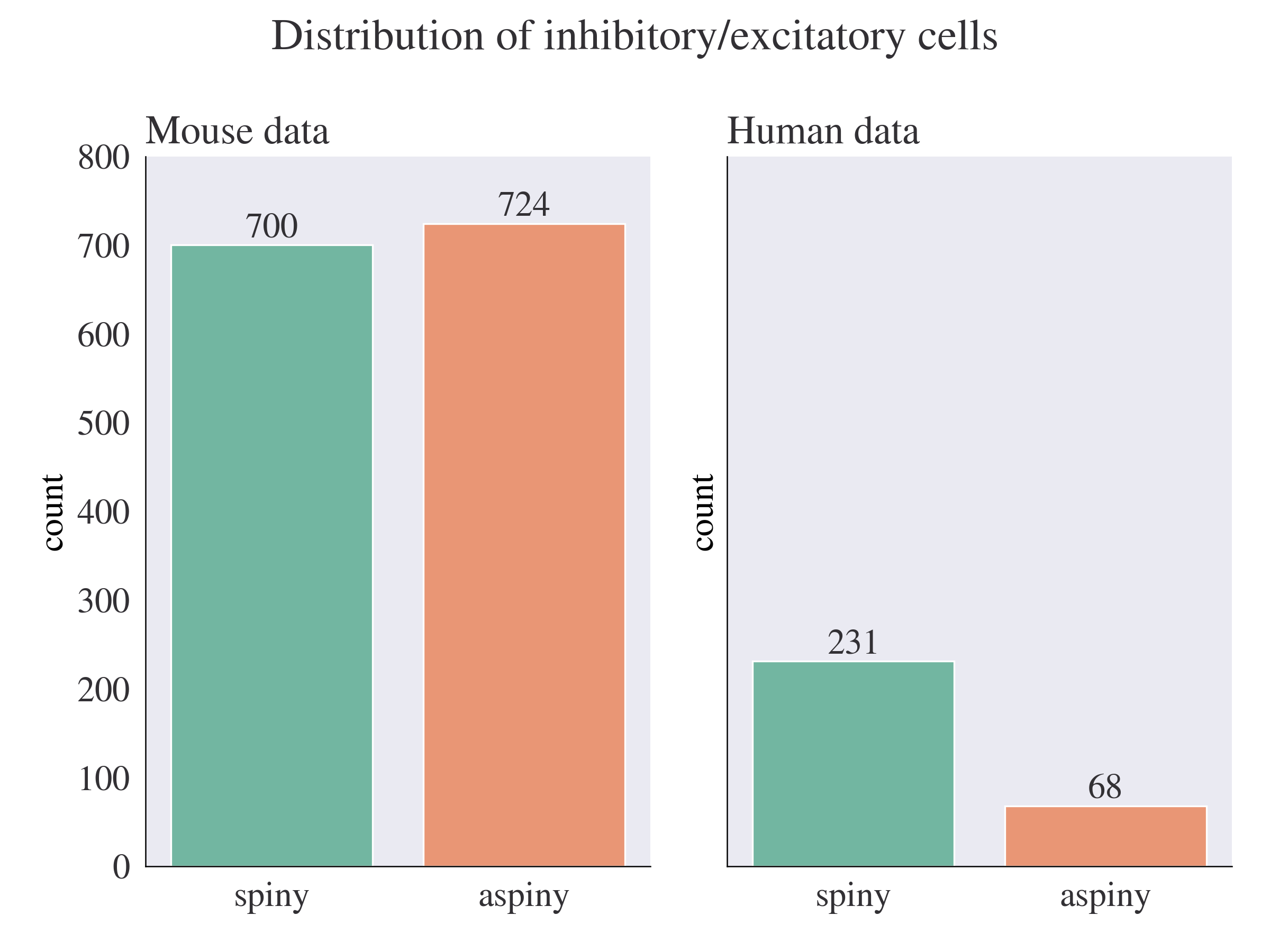}}
\subfigure{\includegraphics[width=0.49\textwidth]{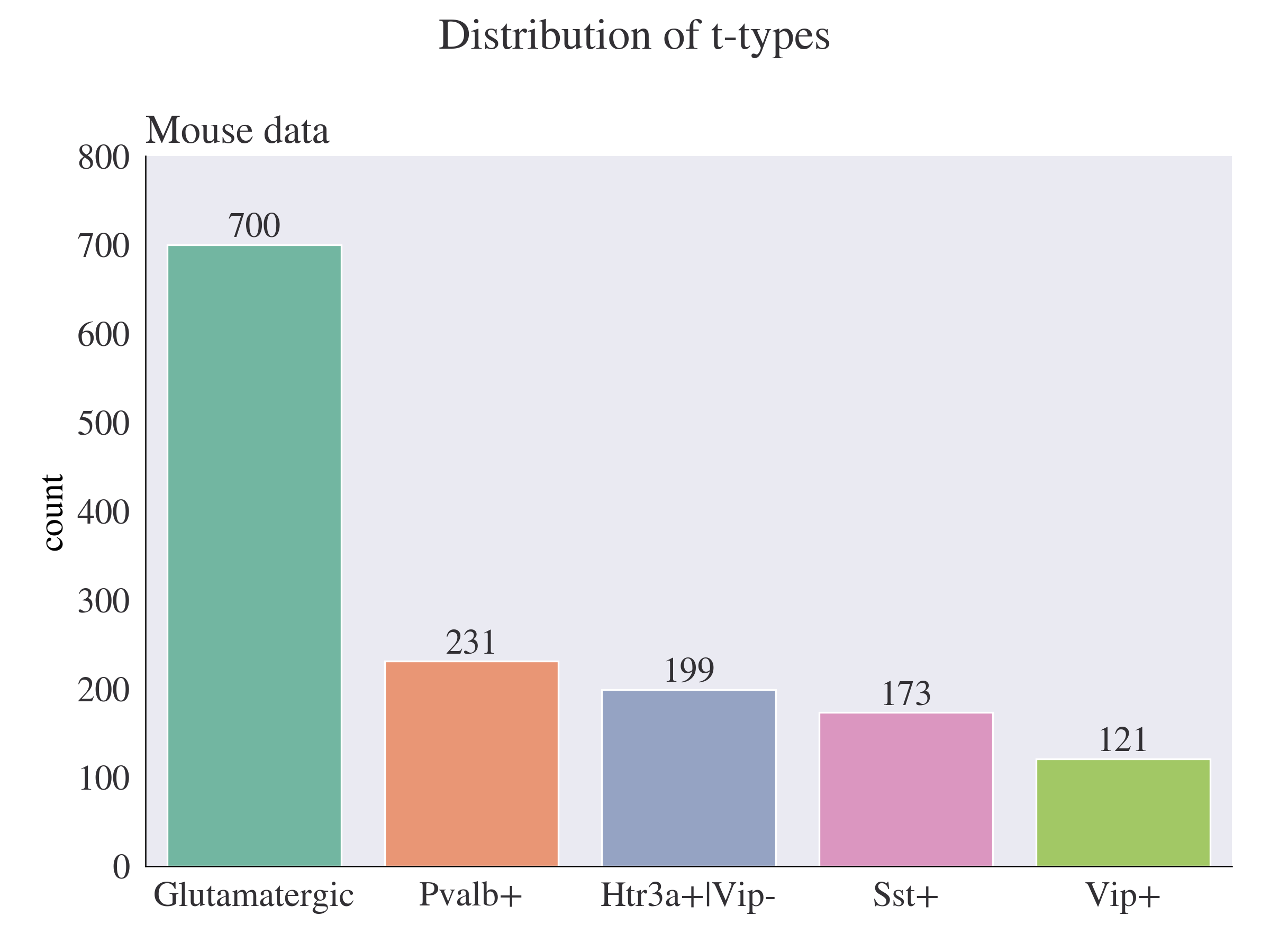}}
\caption{(Left) Distribution of dendrite type in mouse vs. human data (Right) Distribution of the defined Cre-line subclasses in mouse data.}
\end{figure}

\pagebreak

\section{Models}\label{models}
Artificial neural networks (ANN) are computing models inspired by biological neural networks. ANNs rely on matrix multiplications followed by nonlinear activation functions to learn complex relations between input and output. ANNs are comprised of artificial neurons which are connected through edges, these edges typically have a weight value that can adjust the strength of the signal at that connection, and the weights are 'learned' through an optimizer such as Stochastic Gradient Descent (SGD) \cite{ruder2016overview}. \par

Over the last decade, numerous neural network architectures have been proposed. In this paper, we focus on fully connected neural networks, also referred to as multi-layer perceptron (MLP), or just a 'neural network' (NN) \cite{krogh2008artificial}. We also use a new type of NN designed for tabular data, namely LSPIN. \par

We focus on two neuronal cell-type classification tasks: 
\begin{itemize}
    \item  Humans and mice dendritic cell type classification (inhibitory/excitatory)
    \item Multi class (Pvalb, Sst, Vip, Htr3a, Glutamatergic) classificaiton in mice. 
\end{itemize}
For the first task, we introduce a domain adaptation component to handle measurements from humans and mice simultaneously. For the second task, we use a NN with a sample-specific feature selection mechanism, namely LSPIN, to reduce model overfitting in low-sample-size data and obtain an interpretable model. In subsection \ref{domain adaptation}, we describe the domain adaptation mechanism, and in subsection \ref{lspin}, we describe the LSPIN model. \par

\subsection{Domain Adversarial Neural Network}\label{domain adaptation}
Mouse neuronal data is acquired from selected brain areas in adult mice. Cells are identified using transgenic mouse lines harboring fluorescent reporters, with drivers that allow enrichment for cell classes based on marker genes. In contrast, human neuronal data is acquired from donated ex vivo brain tissues analyzed from neurosurgical and postmortem sources and is available thanks to the generosity of tissue donors. Thus, human neuronal data is challenging to obtain and less abundant than data from mice (1920 mouse samples vs. 413 human samples). \par

We aim to design a model that can classify human neuronal types, yet this is difficult due to the scarcity of human samples. To deal with this issue, we use both mouse and human data to classify human samples better. This is possible because mouse and human neurons are similar (both come from mammalian brain tissues). Nonetheless, standard machine learning models typically underperform in such a setting since there may be a domain shift between samples from both distributions. To overcome this limitation, we use a domain adaptation scheme designed to attenuate the influence of domain shift. \par

We consider \(X \in \mathbb{R}^{D}\) the input space, and \(Y \in \{0, 1\}\) the output space, where 0 is an excitatory cell, and 1 is an inhibitory cell.
We define \(S\) to be the source distribution over \(X \times Y\), and \(D_S\) to be the marginal distribution such that \(S = \{(\boldsymbol{x}_i, y_i)\}_{i=1}^n \sim D_S\). 
We define \(T\) to be the target distribution over \(X \times Y\), and \(D_T\) to be the marginal distribution such that \(T = \{(\boldsymbol{x}_i, y_i)\}_{i={n+1}}^{N} \sim D_T\).
Where $n$ is the number of source samples, and $N$ is the number of all samples.
Our goal is to define a classifier \(\eta: X \rightarrow{} Y\) to which the target risk function is low:
\begin{align}\label{eq:loss}
    R_{D_T}(n) = \underset{\{\boldsymbol{x}, y\} \sim D_T}{Pr} (\eta(\boldsymbol{x}) \neq y),
\end{align} \par
\noindent while maintaining a low source risk as well. Since there may be a shift between $D_S$ and $D_T$ training, a naive model based on Eq. \ref{eq:loss} can be biased towards the more abundant domain ($D_S$). To alleviate such bias Ganin, Yaroslav, et al. \cite{ganin2016domain} introduced a technique called 'domain-adversarial training of neural networks (DANN), that combines both representation learning (i.e., deep feature learning) and unsupervised domain adaptation in an end-to-end training process.
DANN jointly optimizes two adversarial losses: 
\begin{enumerate}
    \item minimizing the loss of a label classifier. 
    \item maximizing the loss of a domain classifier. 
\end{enumerate} \par 

Training both losses can be considered a form of adversarial neural network regularization. On the one hand, the network needs to classify the data into the correct labels. On the other hand, the predictions made by the network must be based on features that cannot discriminate between the source domain and the target domain.
In our setting, mouse cells are considered the source distribution and are more abundant (since it is easier to obtain neurons from the rat brain than the human brain), and the human cells serve as the target distribution. \par

\begin{figure}[ht]
\centering
\includegraphics[width=1\textwidth]{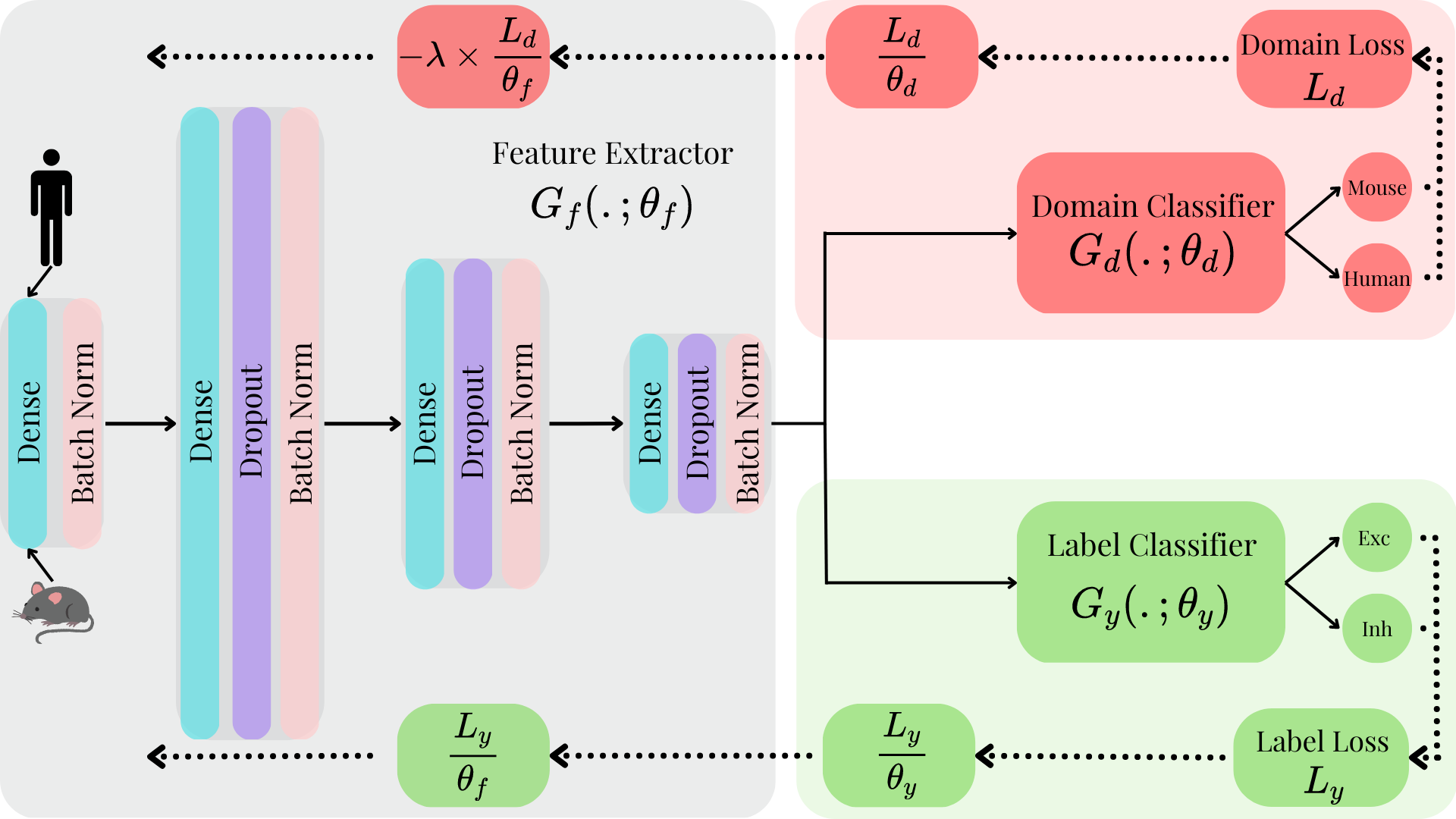}
\caption{The architecture of DANN. During forward propagation (solid-line), input data may come from humans or mice. The feature extractor (gray) with optimized weights \(\theta_f\) outputs to both the domain classifier (red) with optimized weights \(\theta_d\) and label classifier (greed) with optimized weights \(\theta_y\). During backpropagation (dotted line), the domain gradient is multiplied by a negative constant, while the label gradient remains positive.}
\label{fig:dann}
\end{figure}

\noindent The prediction loss and domain loss are respectively defined as
\begin{align*}
    L_y^i(\theta_f, \theta_y) = L_y(G_y(G_f(\boldsymbol{x}_i;\theta_f);\theta_y),y_i),   \\ 
    L_d^i(\theta_f, \theta_d) = L_d(G_d(G_f(\boldsymbol{x}_i;\theta_f);\theta_d),d_i).
\end{align*}
Where \(\theta_f, \theta_y, \theta_d\) are the parameters of the feature extractor, label classifier, and domain classifier, respectively, and $d_i$ is the domain label of sample $i$ as illustrated in Figure \ref{fig:dann}. \par

\noindent Training the model consists of optimizing: 
\begin{equation*}
    E(\theta_f, \theta_y, \theta_d) = \frac{1}{N} \sum_{n=1}^{N} L_y^i(\theta_f, \theta_y) - \frac{\lambda}{N} \sum_{n=1}^{N} L_d^i(\theta_f, \theta_d), 
\end{equation*}\par 

\noindent by finding the saddle point \(\hat{\theta}_f, \hat{\theta}_y, \hat{\theta}_d\) such that: 
\begin{align}\label{eq:minmax_1}
    (\hat{\theta}_f, \hat{\theta}_y) = \arg\min_{\theta_f, \theta_y} E(\theta_f, \theta_y, \hat{\theta}_d), \\
    \hat{\theta}_d = \arg\max_{\theta_d} E(\hat{\theta_f}, \hat{\theta_y}, \theta_d). \nonumber
\end{align} 

\noindent To optimize over Eq. \ref{eq:minmax_1}, we can use gradient descent, which relies on the following update rules:
\begin{align*}
    \theta_f \leftarrow{} \theta_f - \mu(\frac{\partial L_y^i}{\partial \theta_f} - \lambda \frac{\partial L_d^i}{\partial \theta_f}), \\
    \theta_y \leftarrow{} \theta_y - \mu \frac{\partial L_y^i}{\partial \theta_y}, \\
    \theta_d \leftarrow{} \theta_d - \mu \lambda \frac{\partial L_d^i}{\partial \theta_d}.
\end{align*}
Where \(\mu\) is the learning rate. \par

Using the aforementioned NN architecture, domain adaptation is achieved by forcing the prediction based on features that cannot discriminate between mouse and human samples.
Final classification decisions are made using discriminative features invariant to the organism's change. We assume that a good representation for cross-domain transfer is one for which an algorithm cannot identify between the two domains \cite{farahani2021brief,rozner2023domain}. \par

 \par
\subsection{Locally Sparse Neural Network}\label{lspin}
Collecting whole-cell current clamp recordings is computationally challenging; for instance, the ACTB contains only 1920 mouse cells and 413 human cells. The low number of samples makes it challenging to train an over-parametrized NN while avoiding overfitting. To address this obstacle, we adopt a recently proposed method for fitting NN models to low sample size data. Specifically, the method is designed to deal with the problem of low sample size data for tabular heterogeneous biological data such as whole-cell current clamp recordings of neurons in various brain areas in mice. We show that by using the proposed method, we achieve state-of-the-art results. Furthermore, the method proposed is an
intrinsically interpretable network for biomedical data, 'Locally Sparse Interpretable Network' – LSPIN \cite{yang2022locally}. We use LSPIN to predict five distinct neuronal types, four from GABAergic neurons and the remainder from Glutamatergic neurons. \par

The model is a locally sparse neural network in which the local sparsity is learned to identify the subset of the most relevant features for each sample. LSPIN includes two neural networks which are trained in tandem: 
\begin{enumerate}
    \item The gating network - predicts the sample-specific sparsity patterns.
    \item The prediction network - classifies the neuron type using the extracted features from Table \ref{table:features}.
\end{enumerate}
By forcing the model to select a subset of the most informative features for each sample, we can reduce overfitting in low-sample size data. Another benefit of this model is that by predicting the most informative features locally, we obtain an interpretation of the model's predictions. \par

Given labeled observations \(\{\boldsymbol{x}^{(i)}, y^{(i)}\}_{i=1}^N\), where \(\boldsymbol{x}^{(i)} \in \mathbb{R}^D\), and \(x_d^{(i)}\) represents the d\textsuperscript{th} feature of the i\textsuperscript{th} sample. We want to learn a global prediction function \(\boldsymbol{f_\theta}\), and a set of parameters \(\{\mu_d^{(i)}\}_{d=1, i=1}^{D, n}\) such that \(\mu_d^{(i)}\) depict the behavior of the local stochastic gates \(z_d^{(i)} \in [0, 1]\) that sparsify (for each instance $i$) the set of features that propagate into in the prediction model \(\boldsymbol{f_\theta}\). Stochastic gates \cite{yamada2020feature} are continuously relaxed Bernoulli variables highly effective for the sparsification of NNs. They were previously used for several applications, including feature selection \cite{shaham2022deep, jana2021support}, sparse Canonical Correlation Analysis \cite{lindenbauml0}, and anomaly detection \cite{lindenbaum2021probabilistic}. \par

Each stochastic gate (for feature $d$ and sample $i$) is defined based on the following threshold function: 
\begin{align*}
    z_d^{(i)} = max(0, min(1, 0.5 + \mu_d^{(i)} + \epsilon_d^{(i)})),
\end{align*} 
where \(\epsilon_d^{(i)} \sim \mathcal{N}(0,\,\sigma^{2})\ \) and \(\sigma\) is fixed to a constant during training, and equals 0 during inference. The sample-specific parameters \(\boldsymbol{\mu}^{(i)} \in \mathbb{R}^D, i = 1,...,N\) are predicted based on the gating network \(\boldsymbol{\psi}\) such that \(\boldsymbol{\mu}^{(i)} = \boldsymbol{\psi}(\boldsymbol{x}^{(i)}|\boldsymbol{\Omega})\), where \(\boldsymbol{\Omega}\) are the weights of the gating network. These weights are learned simultaneously with the weights of the prediction network by minimizing the following loss: 
\begin{align*}
    \mathbb{E}[\mathcal{L}(\boldsymbol{f_\theta}(\boldsymbol{x}^{(i)} \odot \boldsymbol{z}^{(i)}), y^{(i)}) + \mathcal{R}(\boldsymbol{z}^{(i)})], 
\end{align*} 
where \(\mathcal{L}\) is a desired loss (e.g. cross-entropy), \(\odot\) represents the Hadamard product (element-wise multiplication), and \(\mathcal{R}(\boldsymbol{z}^{(i)})\) is a regularizer term defined as: 
\begin{align*}
    \mathcal{R}(\boldsymbol{z}^{(i)}) = \lambda_1||\boldsymbol{z}^{(i)}||_0 + \lambda_2 \sum_j K_{i,j}||\boldsymbol{z}^{(i)} - \boldsymbol{z}^{(j)}||_2^2,
\end{align*}
where \(K_{i, j} \geq 0\) is a user-defined kernel (e.g., radial basis function).

\begin{figure}[ht]
\centering
\includegraphics[width=1\textwidth]{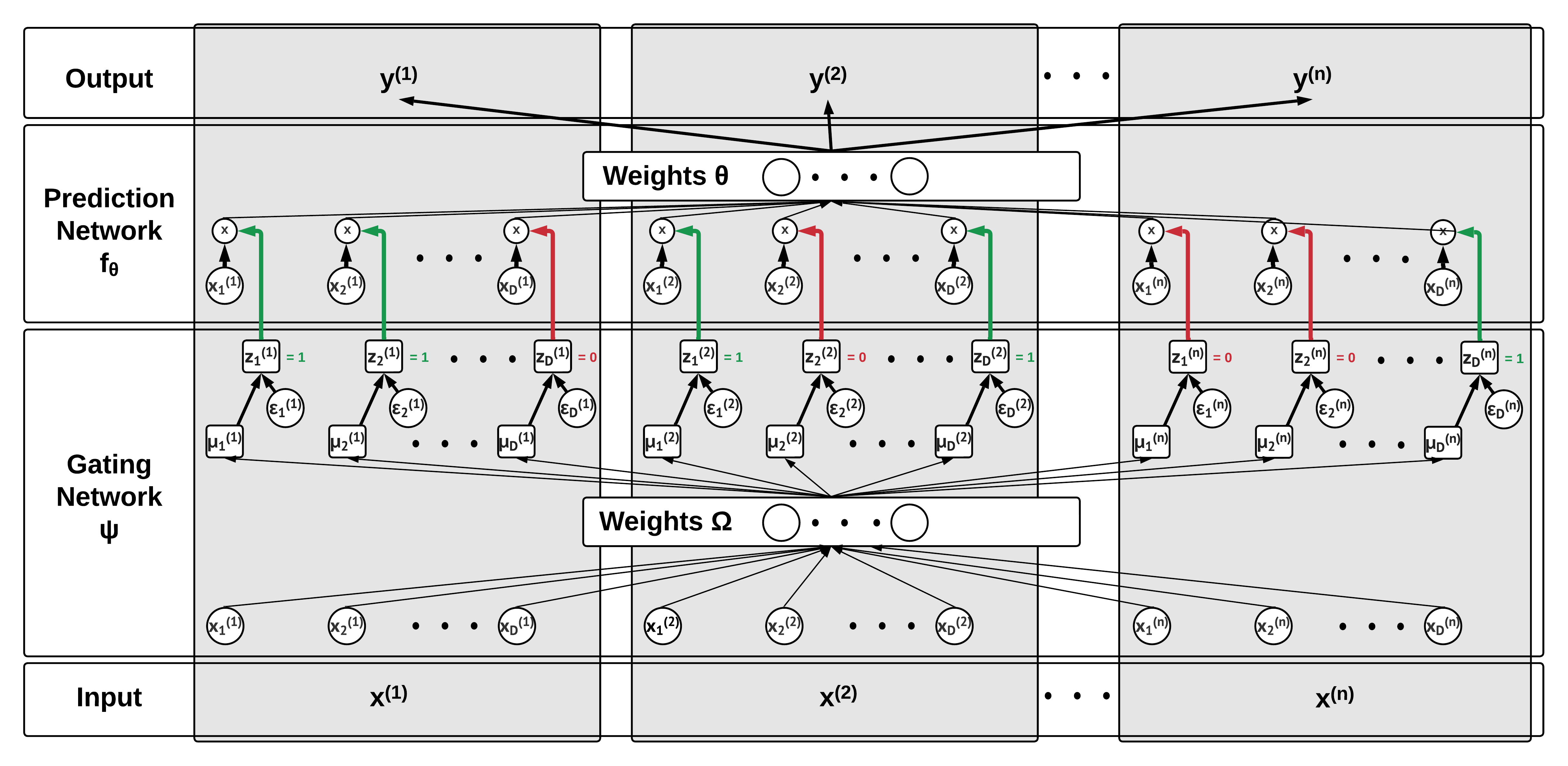}
\caption{The architecture of Locally SParse Interpretable Network. The data \({\boldsymbol{x}^{(i)} = [x_1^{(i)}, x_2^{(i)},..., x_D^{(i)}]}_{i=1}^n\) is fed simultaneously to a gating network \(\boldsymbol{\Psi}\) and to a prediction network \(\boldsymbol{f_\theta}\). The gating model learns to sparsify the set of features propagating to the prediction model, leading to
sample-specific (local) sparsification. Therefore, it can handle extreme cases of Low-Sample-Size data and lead to interpretable predictions.} 
\label{fig3}
\end{figure} \par

\pagebreak

\section{Results}\label{results}
\subsection{Domain Adaptation Prediction Performance}\label{broad type classification}
\begin{figure}[ht]
\centering
\subfigure{\includegraphics[width=0.49\textwidth]{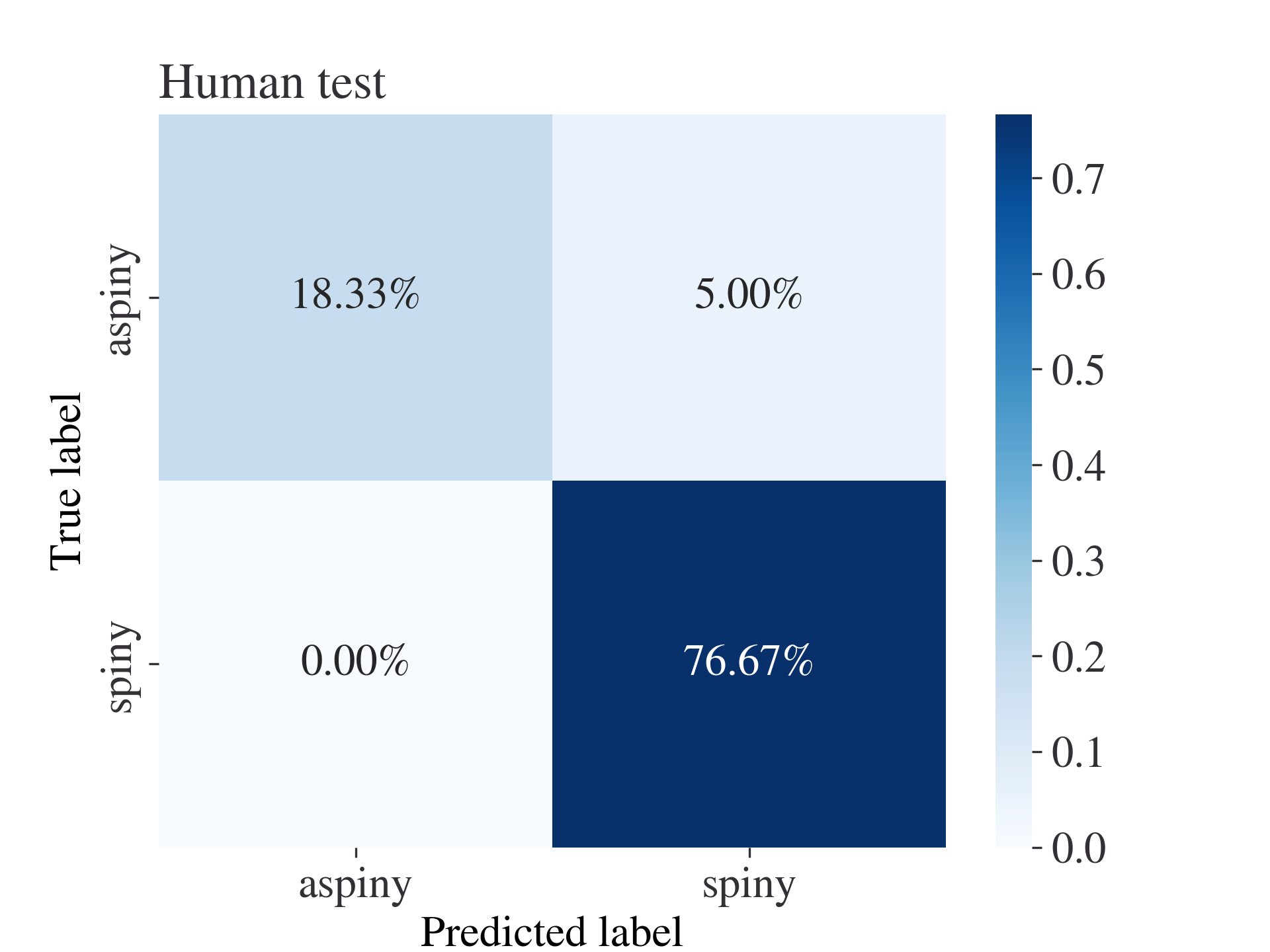}}
\subfigure{\includegraphics[width=0.49\textwidth]{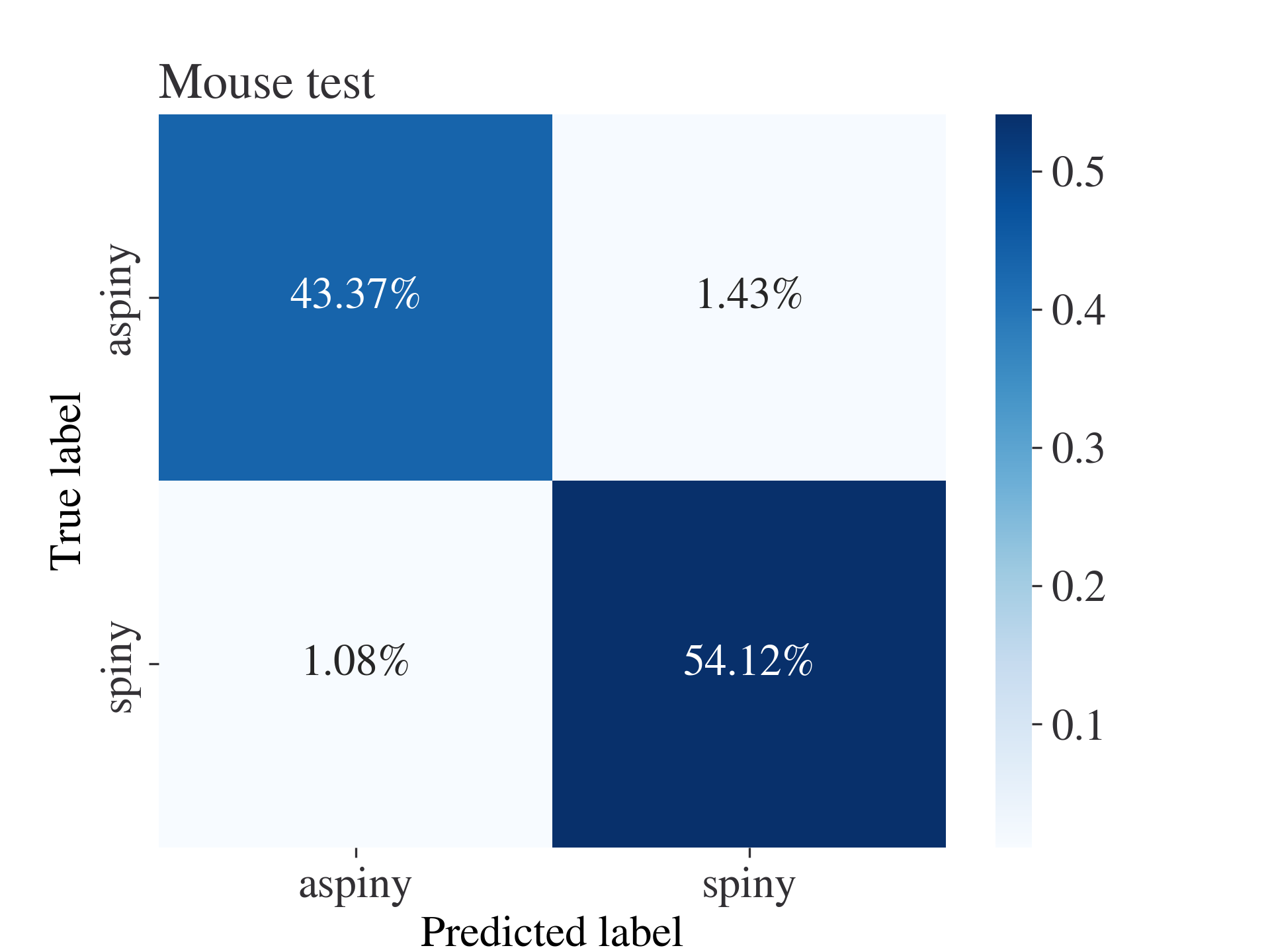}}
\caption{(left) Human classification confusion matrix. (right) Mouse classification confusion matrix. The vertical axis represents the actual class, while the horizontal axis represents the predicted class. The same neural network accurately predicts data from humans and mice.}
\label{fig:dann_results}
\end{figure}

We optimized the DANN model on 1378 training samples of cells from both humans and mice. We split the training data to 1171 samples for training and 207 samples for validation. We also kept 60 human and 285 mouse cells for testing the model. The performance of the domain adaptation task in which neurons are classified based on the broad types (excitatory vs. inhibitory) is shown in Figure \ref{fig:dann_results}. Using the method, we show that the model generalizes to both the human and mouse domains while providing excellent classification results in accuracy, F1 score, precision, and recall. Furthermore, using DANN, we show that whole-cell current clamp recordings of neurons in mouse brains are similar to those of the human brain; moreover, we show that it is possible to adapt to samples from the human domain using the DANN method, providing a gateway in neuronal cell type classification of neurons from the human brain using samples from mouse brains, which are more abundant. \par

 \par

The performance of the DANN model was evaluated using the accuracy, f1 score, precision, and recall metrics. The evaluation results are shown in Table \ref{table:dann_eval}. We show that the method classifies neurons to their broad types with 95.0\% accuracy in human samples and with 97.4\% accuracy in mouse samples using the same model weights, delivering a model that generalizes to both the human and mouse domains and classifies samples from both domains simultaneously.

\begin{center}
\begin{tabular}{|c|c|c|c|c|}
 \hline
  & Accuracy & F1 & Precision & Recall \\ \hline
 Human & 0.950 & 0.968 & 1.000 & 0.938 \\ 
 Mouse & 0.974 & 0.977 & 0.980 & 0.974 \\ 
 \hline
\end{tabular}
\captionof{table}{DANN evaluation metrics for mouse and human data.}
\label{table:dann_eval}
\end{center}
\subsection{Cre Based Sub Classes Prediction Performance}\label{subclasses classification}
\begin{figure}[H]
\centering
\includegraphics[width=0.8\textwidth]{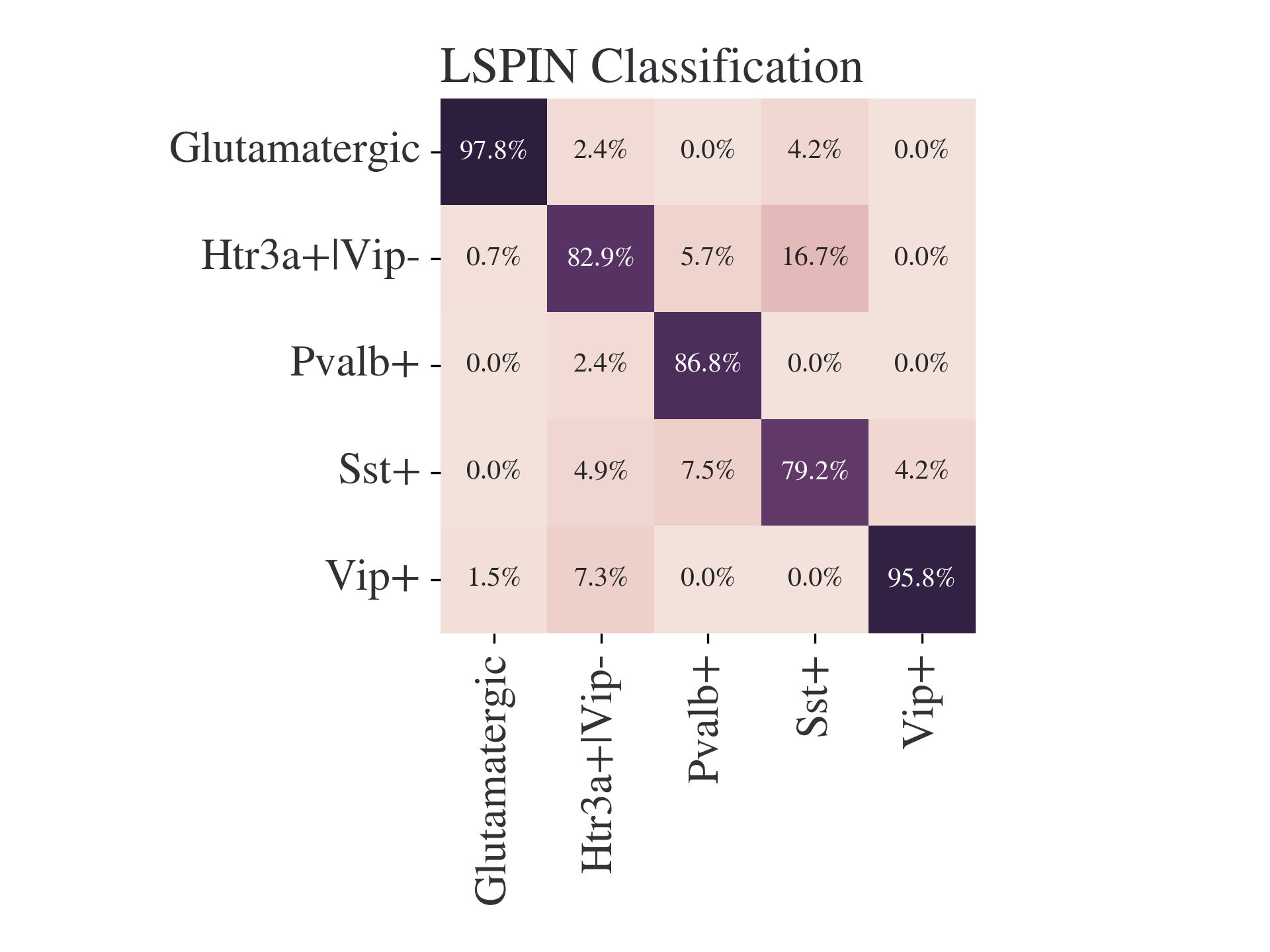}
\caption{Confusion matrix of the LSPIN model. The vertical axis represents the true class, while the horizontal axis represents the predicted class. Glutamatergic neurons are identified 97.8\% of the time correctly, Htr3a neurons are identified 82.9\% of the time, Pvalb 86.8\% of the time, Sst 79.2\% of the time, and Vip neurons are identified 95.8\% of times correctly. Interestingly, Sst neurons are most likely to be confused with Htr3a neurons, suggesting that the two subtypes have similar electrophysiological properties.} 
\label{fig:lspin_results}
\end{figure}

We trained LSPIN on 1105 samples from mice and tested it on 277 (42 samples with NaN values were excluded). The prediction network included two hidden dense layers of size 40 and 20 (with an input layer of size 41 and an output layer of size 5). The gating network was assembled from 3 layers, each containing 50 neurons. Tanh was used as the activation function for both the gating and prediction networks. \(\lambda_1 = 0.01047, \lambda_2 = 0, \sigma = 0.5\), the network was trained for 1000 epochs with a learning rate of 0.0599 and SGD optimizer. The model achieved an accuracy score of 0.916, an F1 score of 0.915, a precision score of 0.917, and a recall score of 0.916. Results of the classification are shown in Figure \ref{fig:lspin_results} \par

\begin{figure}[H]
\centering
\subfigure{\includegraphics[width=0.46\textwidth, valign=t]{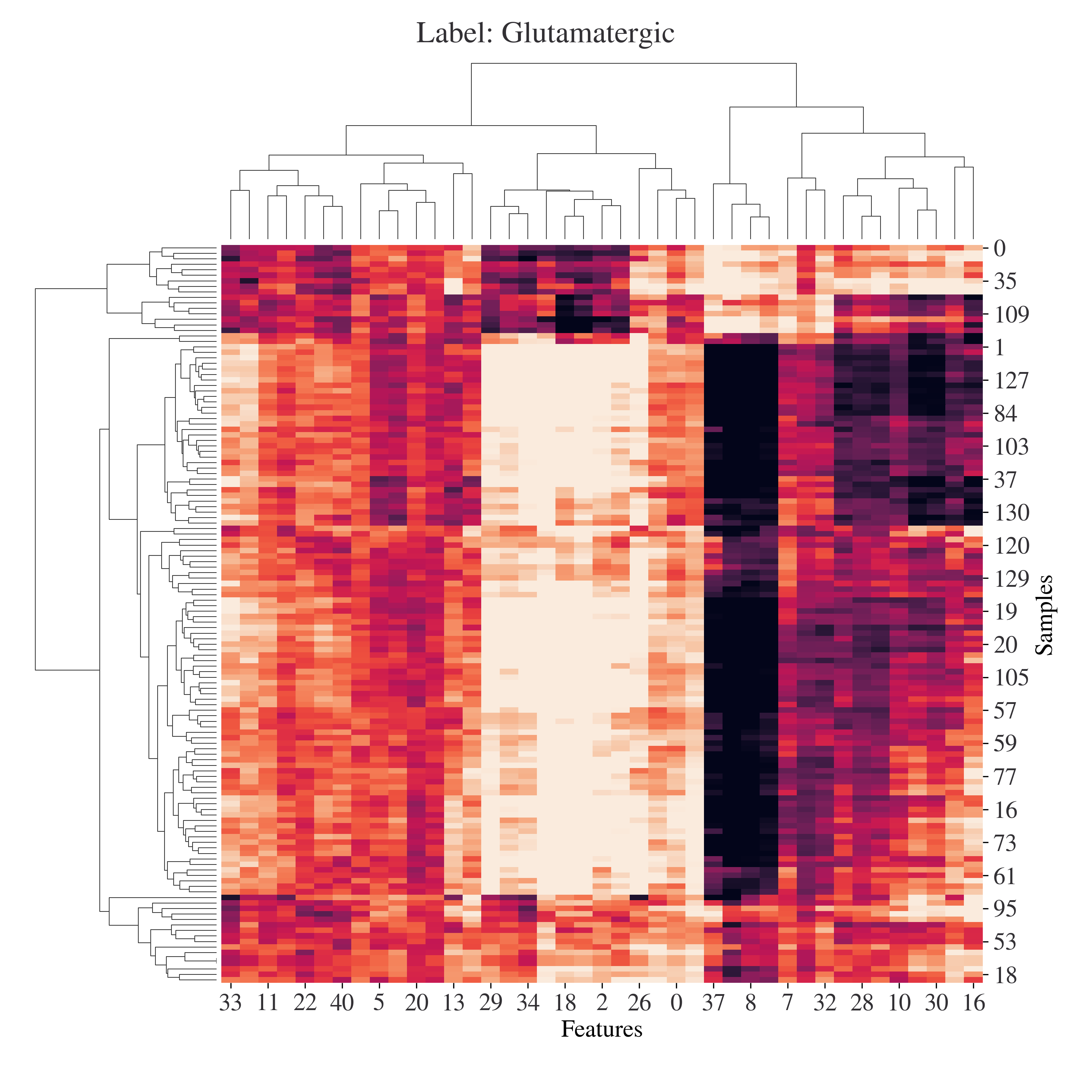}}
\subfigure{\includegraphics[width=0.46\textwidth, valign=t]{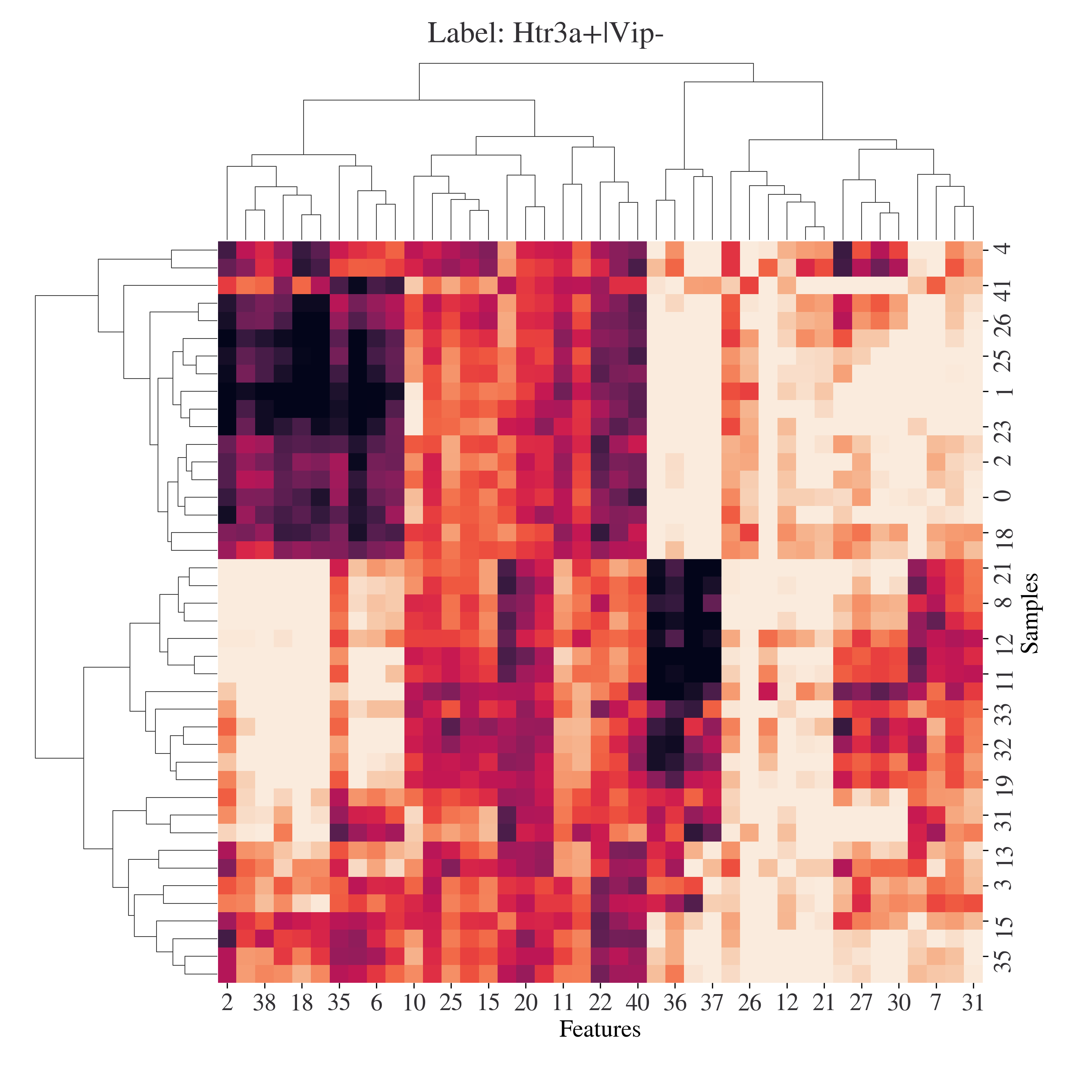}}
\subfigure{\includegraphics[width=0.46\textwidth, valign=t]{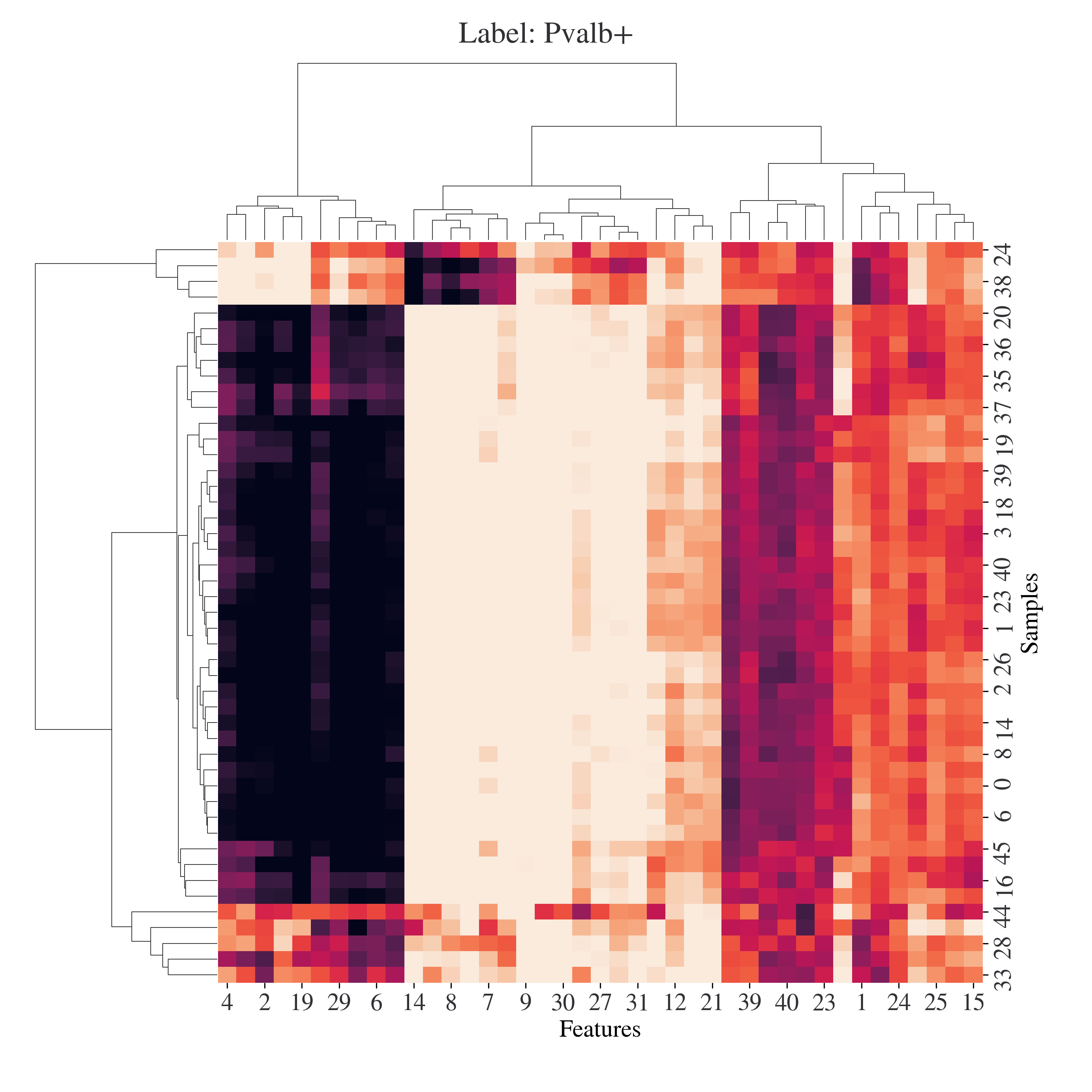}}
\subfigure{\includegraphics[width=0.46\textwidth, valign=t]{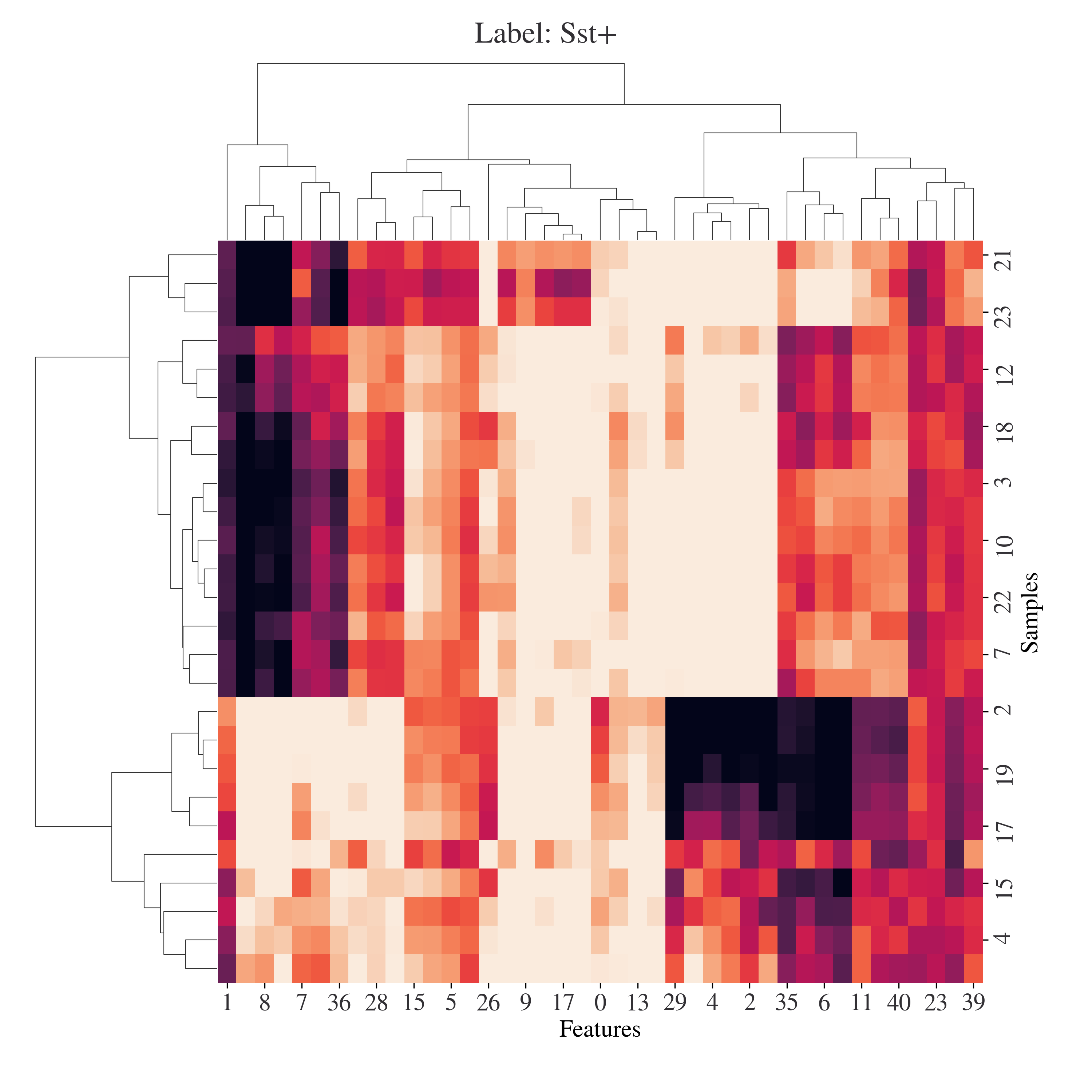}}
\subfigure{\includegraphics[width=0.46\textwidth, valign=t]{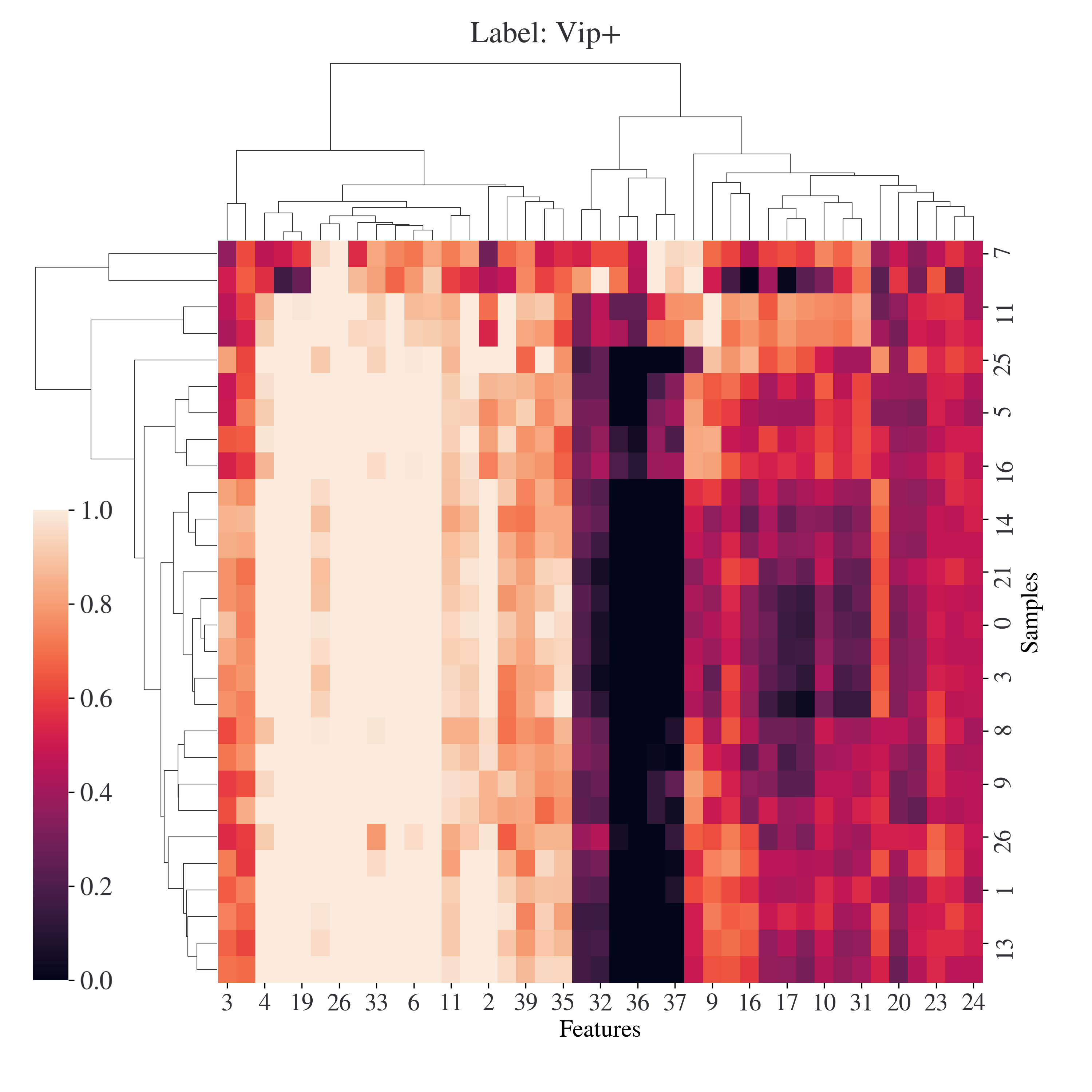}}
\caption{Hierarchical clustering of the gate matrix. Features that are completely muted are displayed in black, while features that stay unchanged are displayed in white. The vertical axis represents the test samples, and the horizontal axis represents the electrophysiological features from Table \ref{table:features} in the appendix.}
\label{fig:gate_matrix}
\end{figure}

Through the gating network and stochastic gates outputs, we can interpret the decisions made by the prediction network, less essential features are muted while important features are not. Figure \ref{fig:gate_matrix} shows the evaluation of features. The model is forced to select a subset of the most informative features identified for each sample. By that, model overfitting is reduced in low-sample-size data, and the selection of relevant features produces an interpretable classification verdict in which relevant and non-relevant features are identified for each neuronal subclass.  \par

 \par

 \par

Furthermore, we trained a Random Forest (RF) classifier, a C-Support Vector Classifier (SVC), and an XGBoost classifier on the same data to compare it with the LSPIN model; the results of the comparison are shown in Table \ref{tab:results}. The results show that the LSPIN model outperforms all other baseline models, achieving 91.6\% accuracy, surpassing XGBoost (the second-best model) by 6.6\%. \par

\begin{center}
\begin{tabular}{|c|c|c|c|c|}
 \hline
  & Accuracy & F1 & Precision & Recall \\ \hline
 RF & 0.824 & 0.726 & 0.774 & 0.702 \\ 
 SVC & 0.837 & 0.749 & 0.789 & 0.731 \\ 
 XGBoost & 0.850 & 0.768 & 0.779 & 0.759 \\
 \textbf{LSPIN} & \textbf{0.916} & \textbf{0.877} & \textbf{0.886} & \textbf{0.873} \\
 \hline
\end{tabular}
\captionof{table}{Comparison of different machine learning models using macro averaging on the same test data. LSPIN outperforms all other optimized models, achieving an accuracy score of 0.916, an F1 score of 0.877, a precision score of 0.886, and a recall score of 0.873.}
\label{tab:results}
\end{center}


\section{Conclusions}\label{conclusions}
We introduce two deep-learning frameworks to identify neuronal electrophysiological types. The first method takes into account both mouse samples and human samples in order to obtain more data. We overcome the domain shift between the two distributions by using a domain adaptation method called 'Domain Adversarial training of Neural Networks.' The second method uses a locally sparse neural network to identify neuronal subtypes while dealing with the issue of low sample size and producing interpretable predictions. \par

In summary, the main contributions of this paper can be concluded as follows:
\begin{itemize}
  \item We introduce a domain adaptive model to classify neurons into excitatory and inhibitory types from both human and mouse samples simultaneously. We show that data from mouse samples can be used to predict neuronal types in data from human samples, which can have significant clinical effects, as human samples are challenging to acquire.
  \item We use the LSPIN model to expand the classification task to neuronal t-type classification based on Cre lines. We show state-of-the-art results in neuronal subtypes classification. The model can also interpret which features are more important to the classification model for each subtype.
\end{itemize} \par

\pagebreak

\bibliographystyle{unsrt}
\bibliography{refs}

\pagebreak

\appendix
\section{Appendix}\label{appendix}

\subsection{Extracted Electrophysiological Features}\label{features}
\begin{table} [ht]
\begin{multicols}{3}
    \begin{enumerate}
        \item threshold v noise
        \item threshold i noise
        \item peak v noise
        \item peak i noise
        \item trough v noise
        \item trough i noise
        \item upstroke ratio noise
        \item upstroke v noise
        \item downstroke ratio noise
        \item downstroke v noise
        \item fast trough v noise
        \item fast trough i noise
        \item width noise
        \item up-down ratio noise
        \item f-i curve slope noise
        \item fast trough v long square
        \item fast trough v ramp
        \item fast trough v short square
        \item input resistance mohm
        \item latency
        \item peak v long square
        \item peak v ramp
        \item peak v short square
        \item ri
        \item sag
        \item seal gohm
        \item tau
        \item threshold i long square
        \item threshold i ramp
        \item threshold i short square
        \item threshold v long square
        \item threshold v ramp
        \item threshold v short square
        \item trough v long square
        \item trough v ramp
        \item trough v short square
        \item up-down ratio long square
        \item up-down ratio ramp
        \item up-down ratio short square
        \item vm for sag
        \item vrest
    \end{enumerate}
\end{multicols}
\caption{Summary of extracted electrophysiological features}
\label{table:features}
\end{table}

\end{document}